\documentclass[sigconf]{acmart}
\AtBeginDocument{%
  }
    
\usepackage[table]{xcolor}
\usepackage{array}
\usepackage{multirow}
\usepackage{algorithm}
\usepackage{algorithmic}
\usepackage{multirow}
\usepackage{pifont}
\usepackage{booktabs}
\usepackage{array}
\usepackage{stfloats}  

\setcopyright{acmlicensed}
\copyrightyear{2026}
\acmYear{2026}
\acmDOI{}
\acmConference[]{}{}{}
\acmISBN{}
\renewcommand\footnotetextcopyrightpermission[1]{}

\begin{document}

\title{CircuitsDNA: Discovering Unconventional Multi-Accuracy Arithmetic Circuits via Evolutionary Synthesis}

\newcommand{\cheng}[1]{{\color{red} #1}}
\newcommand{\junyi}[1]{{\color{blue} #1}}


\author{Ruichen Qi\textsuperscript{1,*}, Junyi Luo\textsuperscript{1,*}, Xinting Jiang\textsuperscript{1}, Quan Cheng\textsuperscript{1}, Gregory Kielian\textsuperscript{2}, Ben Laurie\textsuperscript{2}, Dennis Sylvester\textsuperscript{3}, and Mehdi Saligane\textsuperscript{1,2}}
\affiliation{%
  \institution{\textsuperscript{1}Brown University \quad
  \textsuperscript{2}Google \quad
  \textsuperscript{3}University of Michigan \quad
  \textsuperscript{*}Equal contribution}
  \country{}
}

\renewcommand{\shortauthors}{Anonymous}


\begin{abstract}


Emerging edge AI workloads increasingly require arithmetic units that can trade computational accuracy for efficiency on demand. However, existing approximate arithmetic circuits are typically fixed-accuracy or rely on predefined structures for runtime configurability. This work introduces \textbf{CircuitsDNA}, an evolutionary framework that automatically evolves accuracy-configurable arithmetic circuits supporting multiple accuracy modes within a single circuit. It integrates three key features: 1) multi-threshold verifiability miter to enforce mode-specific accuracy requirements, 2) resource-limited verifiability-driven search to reduce verification overhead without sacrificing correctness, enabling efficient exploration of large circuit design, and 3) feedback-driven adaptive mutation to prioritize effective structural modifications and accelerate search convergence. Experimental results show that the 8-bit multiplier variants synthesized in 28-nm CMOS reduce the area–power product by up to 56\% on INT8 DNN workload and 93\% under exhaustive activity, compared with an exact 8-bit multiplier. Across CNNs and DeiTs, the accuracy loss relative to FP32 remains below 2\% after fine-tuning under worst-case error (WCE) budgets of $\leq$ 1\%. CircuitsDNA eliminates all search stalls observed in conventional methods across 8/12/16-bit multipliers, while adaptive mutation provides up to 1.33$\times$ faster convergence than its non-adaptive counterpart.

\end{abstract}



\maketitle
\pagestyle{plain}  

\setlength{\textfloatsep}{6pt plus 2pt minus 2pt}
\setlength{\floatsep}{5pt plus 2pt minus 2pt}
\setlength{\intextsep}{6pt plus 2pt minus 2pt}
\setlength{\dbltextfloatsep}{7pt plus 2pt minus 2pt}
\setlength{\dblfloatsep}{5pt plus 2pt minus 2pt}
\setlength{\abovedisplayskip}{4pt plus 2pt minus 2pt}
\setlength{\belowdisplayskip}{4pt plus 2pt minus 2pt}
\setlength{\abovedisplayshortskip}{2pt plus 1pt}
\setlength{\belowdisplayshortskip}{3pt plus 1pt}

\begin{figure}[t]
\centering
\includegraphics[width=\columnwidth]{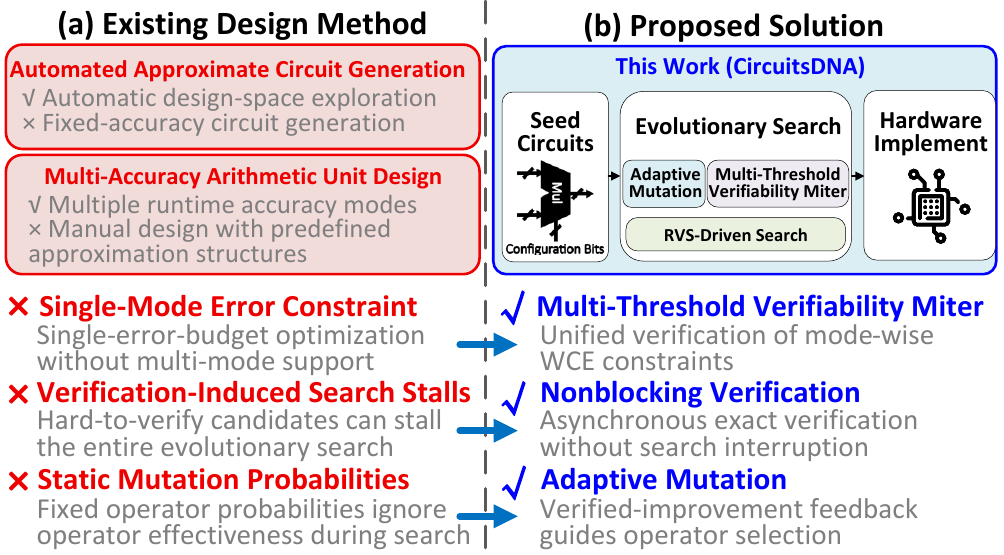}
\vspace{-20pt}
\caption{Motivation and overview of CircuitsDNA.}
\label{fig:comparison}
\end{figure}

\section{Introduction}

Deep neural networks (DNNs) perform billions or even trillions of multiply-accumulate (MAC) operations, leading to high hardware costs \cite{3711683}. Approximate computing (AC) reduces these costs by exploiting the error tolerance of AI workloads~\cite{mittal2016survey,11559576}. However, conventional approximate circuits are typically optimized for a single fixed accuracy--efficiency operating point. This limitation is problematic because error tolerance varies across workloads, layers, and channels~\cite{venkataramani2020efficient,8942068,9643539}. Accommodating such variability requires a single arithmetic circuit to support multiple error levels at runtime while satisfying the error constraint of each mode. Existing automated approaches, however, primarily target fixed-accuracy circuits \cite{8203807}, while runtime-configurable synthesis typically relies on predefined decomposition or approximation structures \cite{alan2020runtime} (see Fig.~\ref{fig:comparison}(a)).

Among these automated approaches, evolutionary algorithms (EAs) \cite{whitley1994genetic, miller2000cartesian}, including genetic algorithms (GAs) and cartesian genetic programming (CGP), are particularly well suited to approximate-circuit design because they can explore discrete and non-differentiable design spaces~\cite{vasicek2014evolutionary}. However, existing EA-based methods~\cite{9586245,jain2022learning,mrazek2017evoapprox8b} mainly optimize individual circuits under fixed accuracy. Extending these methods to synthesize a single circuit with multiple runtime accuracy modes introduces three challenges: 1) a single netlist must simultaneously satisfy multiple mode-specific error constraints; 2) the cost of repeatedly verifying these constraints increases rapidly with circuit size. Invalid candidates can often be rejected quickly, whereas verifying that a candidate satisfies tight error bounds may take substantially longer; and 3) conventional mutation schemes use static, often uniform, operator-selection probabilities throughout the search. However, the effectiveness of different structural modifications varies with the circuit structure and search progress. Fixed probabilities may therefore waste evaluations on ineffective operators and slow convergence.

To address these challenges, we propose \textbf{CircuitsDNA}, an evolutionary synthesis framework that automatically generates accuracy-configurable arithmetic circuits by combining multi-constraint evolution, shared-logic exploration, nonstalling verification, and adaptive mutation (see Fig.~\ref{fig:comparison}(b)). The main contributions are as follows:

\begin{itemize}
\item \textbf{Multi-Threshold Verifiability Miter.} Accuracy-configurable circuit generation requires a single hardware instance to satisfy multiple mode-specific error thresholds. Our unified miter encodes these constraints into one formal-verification instance, enabling efficient verification of runtime-selectable accuracy modes while avoiding separate mode-specific verification and hardware implementations.

\item \textbf{Resource-Limited Verifiability-Driven Search (RVS).} 
We propose RVS to prevent search stalls on hard-to-verify candidates from stalling the overall design-space exploration (DSE). By bounding boolean satisfiability (SAT) verification time and asynchronously handling candidates that exceed the allotted budget, RVS decouples costly verification from the main search, enabling efficient exploration while preserving exact correctness guarantees for all generated circuits.

\item \textbf{Feedback-Driven Adaptive Mutation.}
The adaptive mutation evaluates the effectiveness of different netlist mutation operations during the search and adjusts their mutation operator probabilities accordingly. By favoring more effective operations and reducing less effective mutations, the strategy allocates mutations more efficiently and improves search convergence speed by up to 33\% with negligible overhead.

\end{itemize}

\begin{figure*}[t]
\centering
\includegraphics[width=\textwidth]{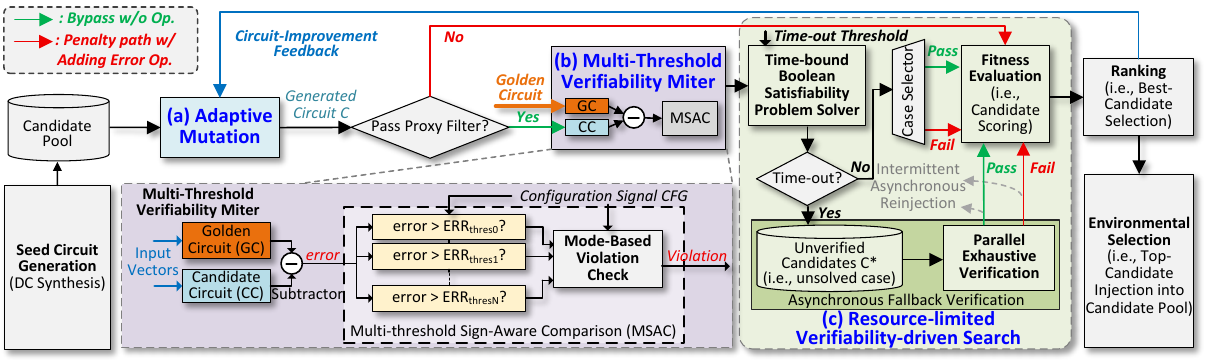}
\caption{Overview of the CircuitsDNA framework with (a) adaptive mutation, (b) multi-threshold verifiability miter, and (c) resource-limited verifiability-driven search.}
\label{circuitsDNA_arch}
\end{figure*}

\section{Related Work}

In AI computing, quantization reduces the precision of weights and activations to lower the computation and memory costs of DNN training and inference~\cite{hubara2018quantized,gupta2015deep}. Mixed-precision computing further assigns different bit widths to operations, layers, or channels according to their accuracy sensitivity~\cite{sharma2018bit,mei2019sub,deng2020model}. However, these methods adapt numerical precision only by selecting from a predefined set of bit widths and therefore explore a limited, discrete design space. They do not exploit the broader PPA-error trade-offs enabled by logic-level approximation within each bit width.

On the other hand, AC expands this design space by relaxing arithmetic correctness at the circuit level. Approximate adders and multipliers have been developed using truncation, logic simplification, compressor redesign, and data-driven optimization~\cite{chen2018exploiting,liu2022approximate,jain2022learning}. Most of these circuits provide only a single fixed accuracy-efficiency operating point and cannot adapt to changing runtime requirements. Accuracy-configurable adders~\cite{6241600} and multipliers for DNN accelerators~\cite{9643491} address this limitation by supporting multiple operating modes within a single circuit. However, their approximation mechanisms are typically designed manually using predefined techniques such as truncation, logic masking, and compression control. Consequently, both their operating modes and circuit structures are restricted by handcrafted templates, limiting structural exploration and the achievable PPA-error trade-offs.

Automated DSE enables broader structural optimization of approximate circuits, while formal verification can enforce strict error bounds during synthesis~\cite{8203807}. Runtime-configurable approximate synthesis has also been investigated using logic gating and relaxation~\cite{alan2020runtime}, as well as truth-table decomposition with configurable approximation blocks~\cite{ma2023ruca}. Nevertheless, these approaches still rely on predefined decomposition or configuration structures. Automatically evolving circuits for multiple runtime-selectable accuracy modes under independent worst-case error constraints remains underexplored. This problem becomes particularly challenging when formal verification is embedded in large-scale structural search, where verification cost can severely limit exploration efficiency.

In summary, prior work has separately explored precision adaptation, runtime configurability, and constrained circuit optimization. CircuitsDNA unifies these directions by jointly evolving a unified circuit structure in which multiple accuracy modes share hardware resources, enabled by multi-constraint evolution, feedback-driven adaptive mutation, and nonstalling formal verification.

\section{CircuitsDNA Framework}

CircuitsDNA (see Fig.~\ref{circuitsDNA_arch}) synthesizes a single circuit supporting multiple runtime accuracy modes. Starting from the candidate pool, adaptive mutation generates a new circuit through structural mutations whose operator probabilities are adjusted based on circuit-improvement feedback (Fig.~\ref{circuitsDNA_arch}(a), Sec.~\ref{subsec:adaptive-mutation}). A lightweight proxy filter then removes candidates unlikely to improve the current solution. Promising candidates enter the multi-threshold verifiability miter, which compares the candidate and golden circuits against all mode-specific error thresholds within a unified verification instance (Fig.~\ref{circuitsDNA_arch}(b), Sec.~\ref{subsec:multi-threshold-miter}). The miter's output is processed by resource-limited verifiability-driven search (RVS), where time-bounded SAT verification handles easy cases while timed-out candidates are transferred to asynchronous exhaustive verification (Fig.~\ref{circuitsDNA_arch}(c), Sec.~\ref{subsec:rvs}). Verification results are incorporated into fitness evaluation, after which candidates are ranked and the top candidates are injected into the pool for subsequent evolution.

\subsection{Adaptive Mutation}
\label{subsec:adaptive-mutation}

\subsubsection{Circuit Representation and Optimization Objective}

Each candidate circuit $C$ is represented as a Directed Acyclic Graph (DAG) whose nodes implement single-output primitive gates from \{AND, NAND, OR, NOR, XOR, XNOR, INV\}. Primary inputs form source nodes, internal nodes represent gates, and directed edges encode signal dependencies. Primary outputs are driven by designated internal nodes. This representation supports structural mutations such as node addition, deletion, and rewiring while preserving valid combinational semantics.

The search minimizes circuit area, defined as the sum of gate-dependent node areas, subject to mode-specific accuracy constraints. For each accuracy mode $c$, the normalized worst-case error (WCE) is

\[
\mathrm{WCE}_c^{\mathrm{norm}}(C)
=
\frac{
\max_{b \in B} \left|y(b)-\hat{y}_c(b)\right|
}{
\max_{b \in B} |y(b)|
}.
\]

Here, $y(b)$ and $\hat{y}_c(b)$ are the outputs of the golden and candidate circuits, respectively, for input $b\in B$. The denominator is the maximum magnitude of the golden output for the corresponding operand width. For an 8-bit two's-complement signed multiplier, for example, it is $|-128\times(-128)|=16384$, so a WCE of $128$ corresponds to a normalized WCE of $0.78\%$.

To minimize circuit area while meeting all mode-specific error constraints, each mode must satisfy its error budget $\varepsilon_c$; otherwise, a large penalty $\Lambda$ is imposed. The genetic algorithm minimizes the fitness $F(C)$:

\[
F(C)=
\begin{cases}
\mathrm{Area}(C), & \forall c,\ \mathrm{WCE}_c^{\mathrm{norm}}(C)\leq \varepsilon_c,\\
\mathrm{Area}(C)+\Lambda, & \text{otherwise}.
\end{cases}
\]

\subsubsection{Structural Mutation Operators}
\label{subsec:mut-ops}

CircuitsDNA uses six structural mutation operators. Node addition (\(\mathcal{M}_{\mathrm{add}}\)) inserts a gate and connects its inputs/outputs; node deletion (\(\mathcal{M}_{\mathrm{del}}\)) removes a node and repairs its connections; gate change (\(\mathcal{M}_{\mathrm{chg}}\)) changes a node's gate type; equivalence merge (\(\mathcal{M}_{\mathrm{merge}}\)) redirects fan-out from redundant logic to an equivalent node; input rewire (\(\mathcal{M}_{\mathrm{rin}}\)) reconnects a node input; and output rewire (\(\mathcal{M}_{\mathrm{rout}}\)) reassigns a primary output.

For each mutation, one operator \(\mathcal{M}_i\) is selected with probability \(w_i\) and applied to the current circuit \(C\) to generate a new candidate \(C'\):
\[
C'=\mathcal{M}_i(C).
\]
The selection probabilities \(w_i\) are adaptively updated during evolution.

\subsubsection{Feedback-Driven Operator Control}
\label{subsec:adaptive-mut}

As shown in Fig.~\ref{circuitsDNA_arch}(a) and Algorithm~\ref{alg:adaptive}, mutation probabilities are updated using a credit-driven Exponentially Weighted Average (EWA). When a verified improvement occurs, all accumulators decay by $(1{-}\alpha)$ and the responsible operator receives an increment of $\alpha$, emphasizing recently effective mutations without updates from uninformative generations. Sampling weights are obtained by clamping each accumulator $e$ to a floor $\delta$ and normalizing, preventing permanent suppression of any operator while retaining adaptive search guidance.

\begin{algorithm}[t]
\small
\caption{Adaptive Mutation with Feedback-Driven Operator Control}
\label{alg:adaptive}
\begin{algorithmic}[1]
\REQUIRE EWA accumulators $\mathbf{e} \in \mathbb{R}^K$, decay rate $\alpha$, floor $\delta$, number of mutation operators $K$
\ENSURE updated sampling weights $\mathbf{w}$

\STATE \textcolor{gray}{// Credit update---triggered only on verified improvement}
\IF{generation $t$ produced a verified improvement via operator $o$}
  \FOR{$i = 1$ \TO $K$}
    \STATE $e_i \gets (1 - \alpha)\, e_i$
  \ENDFOR
  \STATE $e_o \gets e_o + \alpha$
\ENDIF

\STATE \textcolor{gray}{// Weight update}
\FOR{$i = 1$ \TO $K$}
    \STATE $w_i \gets \max(e_i,\; \delta)$
\ENDFOR
\STATE $w_i \gets w_i \;/\; \textstyle\sum_{j=1}^{K} w_j$

\RETURN $\mathbf{w}$
\end{algorithmic}
\end{algorithm}

\subsection{Multi-Threshold Verifiability Miter}
\label{subsec:multi-threshold-miter}

\subsubsection{Proxy-Based Candidate Filtering}

Before formal verification, a lightweight proxy filter removes candidates unlikely to improve the current solution. In our implementation, circuit area serves as the proxy: only candidates improving upon the current best area proceed to the verifiability miter. Candidates failing the filter bypass formal verification and enter fitness evaluation with a penalty. This concentrates verification effort on candidates with improvement potential.

\subsubsection{Unified Multi-Threshold Verification}

Candidates passing the proxy filter enter the multi-threshold verifiability miter in Fig.~\ref{circuitsDNA_arch}(b). A bit-level two's-complement subtractor computes the error between the candidate circuit (CC) and golden circuit (GC). The multi-threshold sign-aware comparison module (MSAC) compares its magnitude against the mode-specific thresholds, while the configuration signal selects the corresponding comparison through the mode-based violation check. The resulting single violation output encodes all mode-specific constraints, allowing all accuracy modes to be verified within one formal-checking instance rather than through separate runs.

\subsection{Resource-Limited Verifiability-Driven Search}
\label{subsec:rvs}

The unified miter is processed by RVS under a bounded verification budget, as shown in Fig.~\ref{circuitsDNA_arch}(c). Cases resolved within the budget immediately enter fitness evaluation, while timed-out candidates are removed from the critical search path and handled by asynchronous fallback verification.

\subsubsection{Time-Bounded SAT Verification}

The miter is translated into a SAT problem and processed by Berkeley ABC's \texttt{iprove} under a strict per-call time limit~\cite{abc}. A violating input can often be found quickly, whereas proving the absence of any violation may require substantially longer. Candidates resolved within the limit are classified by the case selector: verified candidates proceed to fitness evaluation, while candidates with counterexamples receive an error penalty during fitness evaluation and are thereby eliminated.

\subsubsection{Asynchronous Fallback Verification}

If SAT verification reaches the time limit without a definitive result, the unresolved candidate is transferred to a separate queue for parallel exhaustive verification. The fallback engine executes asynchronously and reinjects its pass/fail results into fitness evaluation after completion, preventing difficult cases from stalling the evolutionary loop. Evaluated candidates are then scored and ranked, and environmental selection injects the top candidates into the candidate pool for subsequent generations.

\begin{table}[]
\centering
\small
\setlength{\tabcolsep}{2.5pt}
\renewcommand{\arraystretch}{0.85}
\caption{Seed netlists and nine evolved variant configurations.}
\vspace{-5pt}
\label{tab:evolved_variants}
\resizebox{1.00\columnwidth}{!}{%
\begin{tabular}{
*{3}{>{\columncolor{gray!10}}c}|
*{5}{>{\columncolor{gray!20}}c}
}
\hline
\hline
\multicolumn{3}{>{\columncolor{gray!10}}c|}{
\textbf{Seed Netlist}
} &
\multicolumn{5}{>{\columncolor{gray!20}}c}{
\textbf{Variant Configuration}
} \\
\hline

& & & &
\multicolumn{4}{>{\columncolor{gray!20}}c}{
\rule{0pt}{2.4ex}\textbf{WCE Thres. (\%)}
} \\
\cline{5-8}

\multirow{-2}{*}{\textbf{Seed}} &
\multirow{-2}{*}{
    \textbf{
        \begin{tabular}[c]{@{}c@{}}
        Multiplicand\\[-1pt]
        (A)
        \end{tabular}
    }
} &
\multirow{-2}{*}{
    \textbf{
        \begin{tabular}[c]{@{}c@{}}
        Multiplier\\[-1pt]
        (B)
        \end{tabular}
    }
} &
\multirow{-2}{*}{\textbf{Variant}} &
\textbf{00${}^{a}$} &
\textbf{01${}^{a}$} &
\textbf{10${}^{a}$} &
\textbf{11${}^{a}$} \\
\hline

& & &
$M_{1,1}$ & 0 & 1 & 2.5 & 5.5 \\

& & &
$M_{1,2}$ & 0.1 & 1 & 2.5 & 5.5 \\

& & &
$M_{1,3}$ & 0.2 & 1 & 2.5 & 5.5 \\

\multirow{-4}{*}{Multiplier 1} &
\multirow{-4}{*}{8/7/6/5-bit} &
\multirow{-4}{*}{8-bit} &
$M_{1,4}$ & 0.5 & 1 & 2.5 & 5.5 \\
\hline

& & &
$M_{2,1}$ & 0 & 2.5 & 5.5 & 12 \\

& & &
$M_{2,2}$ & 0.1 & 2.5 & 5.5 & 12 \\

& & &
$M_{2,3}$ & 0.2 & 2.5 & 5.5 & 12 \\

& & &
$M_{2,4}$ & 0.5 & 2.5 & 5.5 & 12 \\

\multirow{-5}{*}{Multiplier 2} &
\multirow{-5}{*}{8/6/5/4-bit} &
\multirow{-5}{*}{8-bit} &
$M_{2,5}$ & 1 & 2.5 & 5.5 & 12 \\
\hline

\multicolumn{8}{>{\columncolor{white}}l}{
${}^{a}$Precision configured via the CFG[1:0] signal
(i.e., A(8/7/6/5-bit)$\times$B(8-bit)).
} \\
\hline\hline

\end{tabular}%
}
\end{table}

\section{Experimental Results}
\label{sec:experiments}

\subsection{Configurable Signed Multiplier Synthesis}

\subsubsection{Experimental Setup}
Table~\ref{tab:evolved_variants} defines nine variants from two seed families. Their modes reduce the effective precision of input $A$, while input $B$ and the output (16-bit) remain fixed. For each variant, we execute 100 independent GA runs on AMD EPYC~7B13 servers; each run takes approximately 20 minutes. The exact, seed, and evolved circuits are synthesized with Synopsys Design Compiler in TSMC 28~nm at 0.9~V and 1~GHz. PrimeTime~PX evaluates power using both all $2^{16}$ input pairs and INT8 ResNet-20 workload. We report $\mathrm{AP}=\mathrm{Area}\times\mathrm{Power}$ normalized to the exact multiplier under the same activity source. Each plotted design is a verified feasible solution selected for minimum AP with nonincreasing power from CFG@[00] to CFG@[11].

\begin{figure}[t]
\centering
\includegraphics[width=\columnwidth]{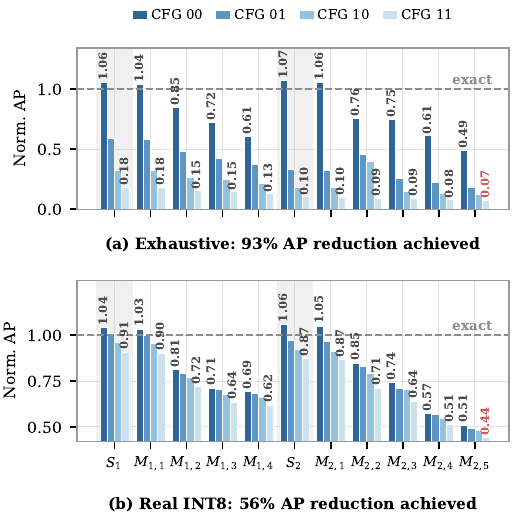}
\caption{Normalized AP under exhaustive inputs and ResNet-20 INT8 workload; S1 and S2 correspond to Seed Multiplier~1 and Multiplier~2 in Table~\ref{tab:evolved_variants}.}
\label{fig:Result_ap}
\vspace{-6pt}
\end{figure}

\subsubsection{PPA-Accuracy Trade-offs}
To evaluate the PPA benefits of configurable approximation, we compare the evolved designs across mode-specific error budgets under exhaustive and real input workload during CNN inference. Figure~\ref{fig:Result_ap} shows that relaxing the mode-wise budgets consistently exposes lower-cost designs. The exact multiplier occupies 207.4~$\mu\mathrm{m}^2$ with a 0.70-ns delay. Under ResNet-20 workload, $\mathrm{M}_{2,5}$ reaches a normalized AP of 0.44 in CFG@[11], a 56\% reduction. Its 151.7-$\mu\mathrm{m}^2$ area is 27\% below the exact baseline, while lower switching supplies the remaining gain. The less aggressive $\mathrm{M}_{1,4}$ still reaches 0.62 normalized AP while restricting CFG@[00] WCE to 0.5\%. Delay also improves by 1.3-8.3\%, showing that AP is not reduced by sacrificing timing.

Exhaustive stimulation yields a larger 93\% maximum AP reduction because it activates switching patterns uncommon in the network workload; the workload-based 56\% result is therefore more deployment-relevant. Since all modes share one circuit, area remains fixed, but changing CFG@[00] to CFG@[11] lowers AP by 10-17\% under real workload. The activity-dependent gap also shows why workload workload are needed for realistic power conclusions.

\begin{figure*}[t]
\centering
\includegraphics[width=\textwidth]{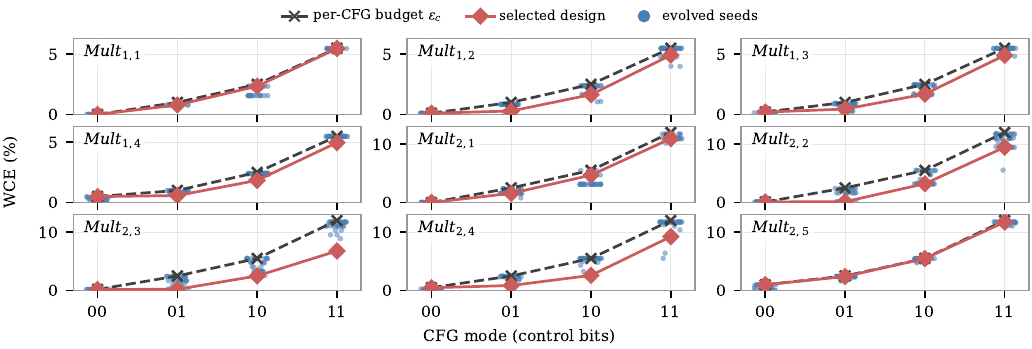}
\vspace{-20pt}
\caption{Mode-wise WCE budgets and achieved WCE across evolved runs.} 
\label{fig:per_cfg_wce}
\end{figure*}

\begin{figure}[t]
\centering
\includegraphics[width=\columnwidth]{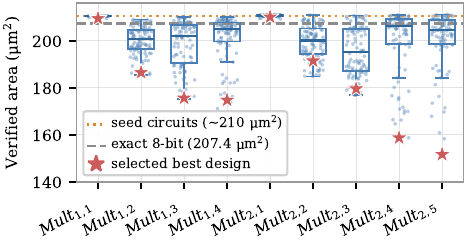}
\vspace{-20pt}
\caption{Final verified area over 100 GA runs per variant.}
\label{area_dist_seeds}
\vspace{-8pt}
\end{figure}

\subsubsection{Constraint Satisfaction and Search Variation}
To verify constraint satisfaction and assess search variation, we examine the achieved WCE and final-area distributions across repeated runs. Figure~\ref{fig:per_cfg_wce} confirms that every selected design satisfies all four WCE constraints. Some modes approach their limits, while others retain slack because WCE is a feasibility constraint, not an objective to maximize. Moreover, all modes compete for the same logic: an area-reducing mutation can be rejected by any one mode, so a final circuit need not lie on all four error boundaries. The result therefore demonstrates simultaneous multi-constraint synthesis rather than separately optimized modes.

Figure~\ref{area_dist_seeds} shows that seven of nine selected variants are smaller than the exact baseline. Only $\mathrm{M}_{1,1}$ and $\mathrm{M}_{2,1}$ remain slightly larger because CFG@[00] is bit-exact while the circuit still contains multi-mode control. Relaxing CFG@[00] progressively lowers the best area. The broad distributions, especially under relaxed budgets, indicate meaningful stochastic variation and justify using repeated runs rather than presenting one seed as representative.

\subsubsection{RVS Ablation}
To evaluate how RVS alleviates verification cost, we compare it with SAT-only search using \emph{verified passes per second}. Both use 32 in-loop ABC workers and the same limit; RVS adds 32 exhaustive-verification workers. Thus, the experiment measures the benefit of a fallback pool at fixed in-loop SAT capacity, not a same-total-core speedup.

\begin{figure}[t]
\centering
\includegraphics[width=\columnwidth]{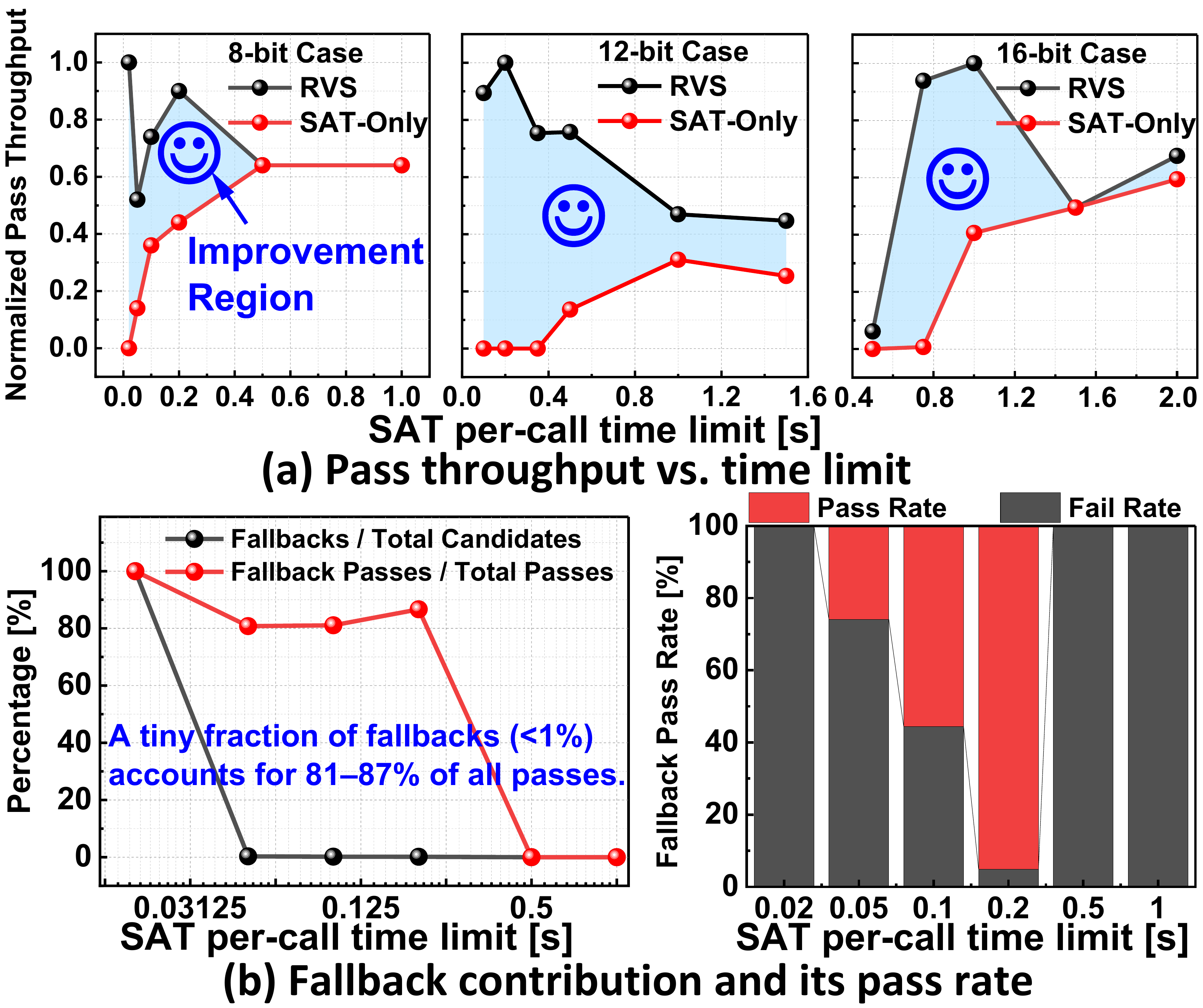}
\vspace{-20pt}
\caption{Effect of the SAT time limit on RVS throughput and fallback effectiveness. 
(a) Normalized pass throughput of RVS and SAT-only verification. 
(b) Fallback frequency, contribution to total passes, and pass rate within fallback candidates.}
\label{fallback_analysis}
\vspace{-1pt}
\end{figure}

Figure~\ref{fallback_analysis}(a) shows that the SAT limit required for useful in-loop evolving increases with width. At tight limits, SAT-only produces no verified passes and remains at its seed, whereas RVS moves unresolved candidates off the critical path and continues searching. At 16 bits, SAT-only passes nothing at 0.5~s while RVS still progresses; the advantage disappears only when the limit becomes long enough for ABC to resolve candidates in-loop. Hence, one timeout cannot simultaneously minimize stalling and preserve proof throughput across widths.

The 8-bit breakdown in Fig.~\ref{fallback_analysis}(b) explains the effect. At 0.05-0.2~s, fewer than 1\% of calls enter the fallback, yet they contribute 81-87\% of verified passes. Timed-out cases are therefore enriched with proof-difficult but feasible candidates. Their fallback pass rate rises from 26\% to 95\% as the SAT limit increases, and the fallback stops firing at 0.5\,s. RVS is consequently self-limiting: nearly inactive when SAT is sufficient, but decisive when rare hard proofs would otherwise stall evolution.

\subsubsection{Adaptive-Mutation Ablation}
To evaluate whether adaptive mutation accelerates convergence without compromising the quality of the evolved circuits, we compare it with uniform selection of the six mutation operators using population size 2, 10 paired seeds, and 200k generations. The EWA rate $\alpha$ is swept from 0.01 to 0.30, and speedup measures the reduction in generations needed to reach the same area target.

\begin{figure}[t]
\centering
\includegraphics[width=\columnwidth]{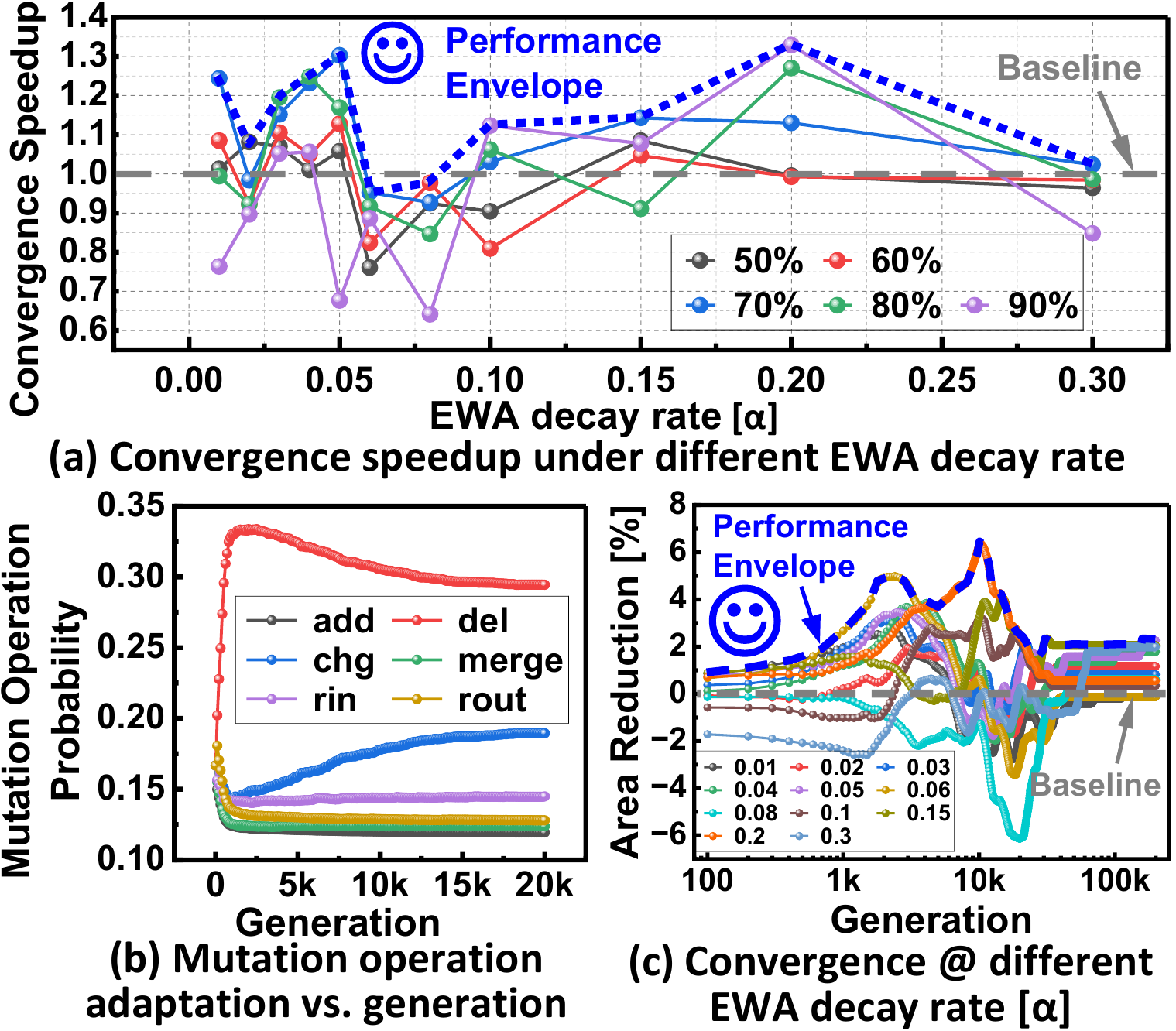}
\vspace{-20pt}
\caption{Impact of the EWA decay rate on adaptive mutation. (a) Convergence speedup across area-improvement targets. (b) Evolution of mutation-operator probabilities. (c) Relative area reduction across EWA decay rates.}
\label{adaptive_baseline}
\vspace{-1pt}
\end{figure}

Figure~\ref{adaptive_baseline}(a) shows that the convergence speedup achieved by our proposed adaptive mutation method depends on the decay rate. Here, each percentage (i.e., 50-90\%) denotes the fraction of the area optimization achieved by the converged result. The longer-memory $\alpha=0.05$ reaches the 50\% target $1.33\times$ faster than fixed mutation, whereas a larger $\alpha=0.20$ gives the largest near-final (90\%) speedup of $1.32\times$. The performance envelope further suggests that dynamically adapting $\alpha$ during search could track the best-performing setting and maximize convergence speedup without sacrificing final area.

The learned probabilities explain this acceleration. Node deletion rises to approximately 0.33 early and remains dominant, while gate change becomes second for local refinement. Consistently, the envelope in Fig.~\ref{adaptive_baseline}(c) shows that different $\alpha$ values provide the best area reduction at different generations, with up to about 2.5\% extra area saving compared to the baseline after convergence. Thus, dynamically adapting $\alpha$ can better exploit the search-dependent effectiveness of mutation operators without compromising search quality.

\begin{table}[t]
\centering
\small
\setlength{\tabcolsep}{4pt}
\renewcommand{\arraystretch}{0.85}
\caption{Top-1 accuracy (\%) after one-epoch fine-tuning with one global CFG mode.}
\vspace{-10pt}
\label{tab:global_cfg}
\begin{tabular}{lccccc}
\toprule
\hline
\textbf{Network (dataset)} & \textbf{FP32 ref.} & \textbf{0.1\%} & \textbf{1\%} & \textbf{2.5\%} & \textbf{5.5\%} \\
\midrule
LeNet-5 (MNIST)        & 99.20 & 99.2 & 99.2 & 99.2 & 99.1 \\
ResNet-20 (CIFAR-10)   & 92.98 & 92.8 & 92.8 & 92.5 & 91.5 \\
ResNet-18 (CIFAR-100)  & 78.29 & 77.4 & 77.2 & 74.5 & n/a \\
ResNet-50 (ImageNet)   & 80.86 & 79.5 & N/A & N/A & N/A \\
\midrule
DeiT-Tiny (CIFAR-100)  & 83.8 & 82.2 & 81.9 & N/A & N/A \\
DeiT-Small (CIFAR-100) & 88.1 & 86.3 & 86.1 & N/A & N/A \\\hline
\bottomrule
\end{tabular}
\vspace{-6pt}
\end{table}

\begin{table*}[!b]
\centering
\caption{Comparison with SOTA approximate-arithmetic and runtime-configurable works.}
\vspace{-10pt}
\label{tab:sota_comparison}
\renewcommand{\arraystretch}{0.9}
\resizebox{\textwidth}{!}{
\begin{tabular}{lccccc}
\toprule
\hline
\multicolumn{1}{c}{\multirow{2}{*}{\textbf{Work}}} &
\multirow{2}{*}{\textbf{Method}} &
\textbf{Runtime Configuration} &
\multirow{2}{*}{\shortstack{\textbf{Automatic}\\\textbf{Generation}}} &
\multirow{2}{*}{\textbf{Error Metric}} &
\multirow{2}{*}{\textbf{DNN Evaluation}} \\[-2pt]
& & \textbf{(single instance)} & & & \\
\midrule

Ceska et al.~\cite{8203807}
& CGP + formal checking
& No
& Yes
& WCAE
& No \\

RUCA~\cite{ma2023ruca}
& BMF + truth-table decomposition
& Yes
& Yes
& QoR error threshold
& No \\

DASALS~\cite{wang2023dasals}
& Differentiable architecture search
& No
& Yes
& MSE
& No \\

ACBAM~\cite{roy2022acbam}
& Configurable broken-array Booth
& Yes
& No
& MRED
& No \\

Rank-based ALS~\cite{ye2025rankals}
& MCTS + rank prediction
& No
& Yes
& ER / NMED
& No \\

IDA-AxM~\cite{buccolini2026ida}
& SMT + distribution-aware constraints
& No
& Yes
& Distribution-weighted MeanAE
& ResNet-8 \\

\textbf{CircuitsDNA}
& \textbf{Multi-constraint evolution + RVS}
& \textbf{Yes}
& \textbf{Yes}
& \textbf{Mode-wise normalized WCE}
& \textbf{LeNet-5, ResNet-18/20/50, DeiT-T/S} \\\hline

\bottomrule
\end{tabular}
}
\end{table*}

\subsection{Neural-Network Evaluation}
\label{sec:cam_nn}

To evaluate end-to-end NN accuracy and energy benefits, we replace each PE multiplier in an output-stationary systolic array with $\mathrm{M}_{1,2}$ for DNN inference. A shared 2-bit register supports offline per-layer selection without changing the datapath. To isolate each operating point, Table~\ref{tab:global_cfg} applies one CFG network-wide. Models receive one fine-tuning epoch; ResNet-50 uses 1\% of ImageNet per epoch~\cite{pytorch_vision_models}. Under real DNN traces, the approximate modes consume $6.8\times10^{-2}$ to $7.7\times10^{-2}$~pJ per multiplication, compared with $8.56\times10^{-2}$~pJ for the exact unit, corresponding to energy savings of 10--21\%.

Table~\ref{tab:global_cfg} shows that error tolerance is workload-dependent. LeNet-5 and ResNet-20 tolerate 5.5\% WCE with accuracy losses of only 0.10 and 1.48 percentage points relative to FP32. ResNet-18 tolerates 1\% WCE with a 1.09-point loss, while ResNet-50 incurs a 1.36-point loss at 0.1\% WCE. DeiT-Tiny and DeiT-Small lose 1.9--2.0 points at 1\% WCE, demonstrating applicability beyond CNNs but a preference for tighter modes. Separate exact-INT8 runs remain within approximately one point of FP32; therefore, the FP32 column is a common reference rather than a direct approximate-versus-exact-INT8 delta. These results motivate workload- or layer-specific selection instead of a universal WCE setting.

\subsection{Bit-Width Scalability}
\begin{figure}[t]
\centerline{\includegraphics[width=\columnwidth]{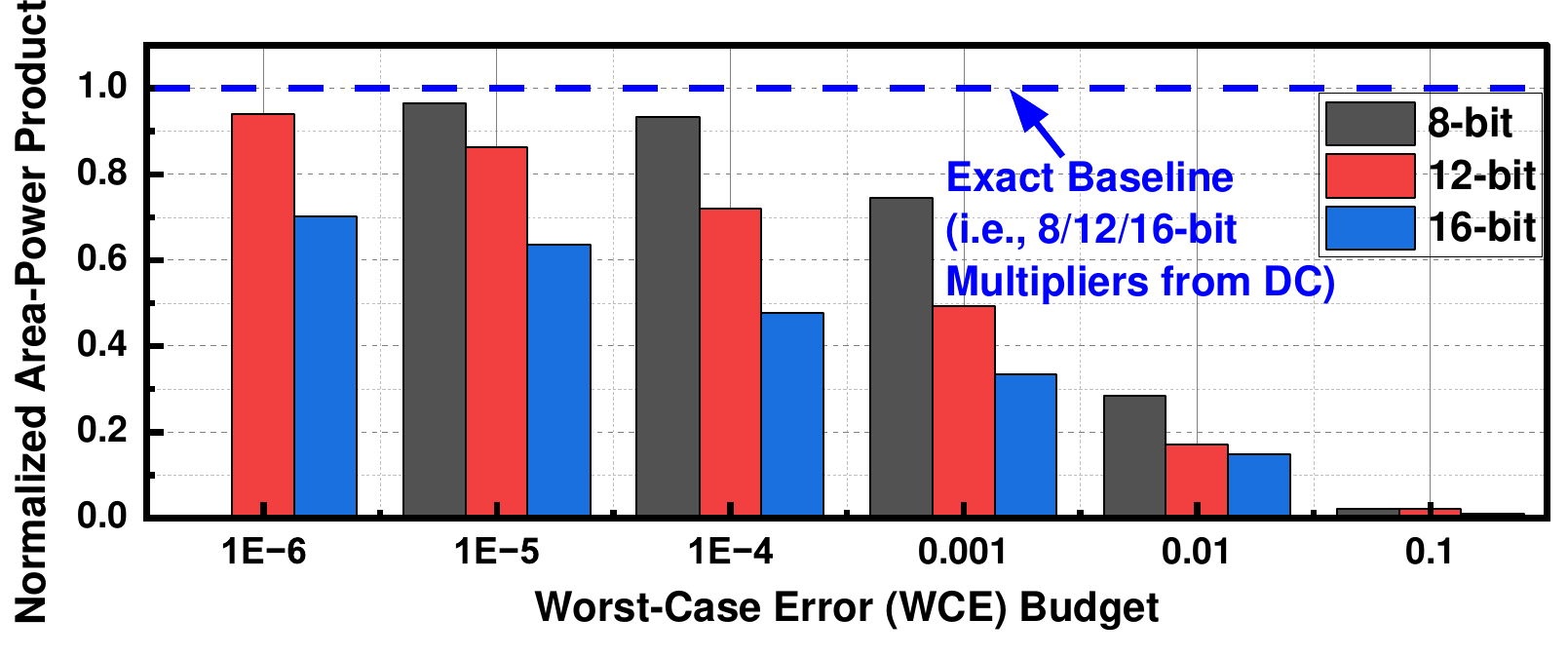}}
\vspace{-10pt}
\caption{Normalized area-power product versus WCE budget.}
\label{fig:scalability}
\end{figure}

To evaluate scalability beyond INT8, we evolve 12- and 16-bit multipliers using 20 and 3 seeds, respectively, with population size 2 and only the highest-accuracy mode. Figure~\ref{fig:scalability} shows the same qualitative trend across widths: relaxing WCE exposes lower-AP circuits. Search completes in approximately 2~hours at 12 bits and 5~hours at 16 bits despite harder verification. Because the 16-bit experiment uses only three seeds and considers only the highest-accuracy mode, it demonstrates feasibility beyond INT8 rather than providing a statistically matched cross-width PPA comparison.

\subsection{Comparison with Existing Work}

Table~\ref{tab:sota_comparison} compares CircuitsDNA with recent automated and runtime-configurable approximate-circuit design methods. Existing automated methods generally generate separate fixed-accuracy circuits, whereas runtime-configurable designs often rely on predefined approximation structures. As these works use different technologies, workloads, and error metrics, the table focuses on capability coverage rather than numerical PPA results. CircuitsDNA jointly evolves a unified netlist with multiple runtime-selectable accuracy modes under independent mode-wise WCE constraints, combining automatic structural generation, single-instance runtime configurability, formal error guarantees, and evaluation across both CNN and transformer workloads.

\section{Conclusion}


This work presented CircuitsDNA, an evolutionary synthesis framework that automatically generates a unified arithmetic circuit with multiple runtime-selectable accuracy modes under independent mode-wise WCE constraints. The generated 8-bit multiplier variants reduce the area-power product by up to 56\% on on ResNet-20 INT8 workload and 93\% under exhaustive activity relative to an exact multiplier. Across the evaluated CNNs and DeiTs, the accuracy loss relative to FP32 remains below 2\% after fine-tuning under worst-case error (WCE) budgets of $\leq 1\%$. RVS eliminates the search stalls observed with synchronous verification across the 8-, 12-, and 16-bit multipliers, while adaptive mutation accelerates convergence by up to 1.33$\times$. These results demonstrate automated synthesis of formally bounded, runtime-configurable arithmetic circuits without relying on predefined approximation structures.

\clearpage
\clearpage
\bibliographystyle{unsrtnat}
\bibliography{sample-base}

\appendix

\end{document}